\documentclass[runningheads]{llncs}
\usepackage{pgfplots}
\pgfplotsset{compat=1.17}

\usepackage{graphicx}
\usepackage{svg}
\usepackage{pgfplots}
\usepackage{amsfonts} 
\usepackage{multirow}
\usepackage[skip=2pt plus1pt, indent=10pt]{parskip}

\newcommand{\cmark}{\ding{51}}%
\newcommand{\xmark}{\ding{55}}%

\usepackage{url}

\usepackage{amssymb}% http://ctan.org/pkg/amssymb
\usepackage{pifont}% http://ctan.org/pkg/pifont

\AtBeginDocument{%
  \providecommand\BibTeX{{%
    \normalfont B\kern-0.5em{\scshape i\kern-0.25em b}\kern-0.8em\TeX}}}
    
\usepackage{algorithm}
\usepackage{algorithmic}
\usepackage{bm}

\usepackage{booktabs}
\usepackage{graphicx}

\usepackage{amsthm}
\usepackage{amsmath,amsfonts}

\usepackage{booktabs}
\usepackage{caption}
\usepackage{textcomp}
\usepackage{subfigure}
\theoremstyle{plain}

\usepackage{todonotes} % self
\newcommand{\hide}[1]{}
\usepackage[normalem]{ulem}

\begin{document}
\setlength{\parskip}{0pt}

%
% --- Author Metadata here ---
% -- Can be completely blank or contain 'commented' information like this...
%\conferenceinfo{WOODSTOCK}{'97 El Paso, Texas USA} % If you happen to know the conference location etc.
%\CopyrightYear{2001} % Allows a non-default  copyright year  to be 'entered' - IF NEED BE.
%\crdata{0-12345-67-8/90/01}  % Allows non-default copyright data to be 'entered' - IF NEED BE.
% --- End of author Metadata ---

\title{Deep Graph Stream SVDD: Anomaly Detection in Cyber-Physical Systems}
\author{Ehtesamul Azim \and
Dongjie Wang \and
Yanjie Fu}
\authorrunning{Ehtesam et al.}
\institute{Department of Computer Science\\ University of Central Florida, Orlando, FL 32826, USA \\
\email{\{azim.ehtesam,wangdongjie\}@knights.ucf.edu, yanjie.fu@ucf.edu}}
\maketitle
\begin{abstract}
% This paper provides a sample of a LaTeX document for final submission
% to Sigkdd Explorations, the official newsletter of ACM Sigkdd. This is
% a modified version of the ACM Proceedings sample file.

% The developers have tried to include every imaginable sort
% of ``bells and whistles", such as a subtitle, footnotes on
% title, subtitle and authors, as well as in the text, and
% every optional component (e.g. Acknowledgements, Additional
% Authors, Appendices), not to mention examples of
% equations, theorems, tables and figures.

% To make best use of this sample document, run it through \LaTeX\
% and BibTeX, and compare this source code with the printed
% output produced by the dvi file.
% \section{Proposal}
% \newline

% dongjie 
% Cyber-physical systems (CPS) have been deployed everywhere and play a significant role in the real world, including smart grids, robotics systems, water treatment networks, etc.
% Due to complex interactions and invoking relations in such systems, they are vulnerable to abnormal system events (e.g., cyberattack, system exception).
% To preserve the stability of CPS and prevent catastrophic failures, considerable research attention has been devoted to anomaly detection based on sensor monitoring data generated by these systems.
Our work focuses on anomaly detection in cyber-physical systems.
Prior literature has three limitations: 
(1) Failing to capture long-delayed patterns in system anomalies;
(2) Ignoring dynamic changes in sensor connections;
(3) The curse of high-dimensional data samples.
These limit the detection performance and usefulness of existing works.
To address them, we propose a new approach called deep graph stream support vector data description (SVDD) for anomaly detection.
Specifically, we first use a transformer to preserve both short and long temporal patterns of monitoring data in temporal embeddings.
Then we cluster these embeddings according to sensor type and utilize them to estimate the change in connectivity between various sensors to construct a new weighted graph.
The temporal embeddings are mapped to the new graph as node attributes to form weighted attributed graph.
We input the graph into a variational graph auto-encoder model to learn  final spatio-temporal representation.
Finally, we learn a hypersphere that encompasses normal embeddings and predict the system status by calculating the distances between the hypersphere and data samples.
Extensive experiments validate the superiority of our model, which improves F1-score by 35.87\%, AUC by 19.32\%, while being 32 times faster than the best baseline at training and inference.

\end{abstract}
% \vspace{-0.7cm}

\section{Introduction}
Cyber-physical systems (CPS) have been deployed everywhere and play a significant role in the real world, including smart grids, robotics systems, water treatment networks, etc.
Due to their complex dependencies and relationships, these systems are vulnerable to abnormal system events (e.g., cyberattacks, system exceptions), which can cause catastrophic failures and expensive costs.
In 2021, hackers infiltrated Florida's water treatment plants and boosted the sodium hydroxide level in the water supply by 100 times of the normal level \cite{florida_cyber}. This may endanger the physical health of all Floridians. 
To maintain stable and safe CPS, considerable research effort has been devoted to effectively detect anomalies in such systems using sensor monitoring data \cite{zhou2020siamese,wang2023hierarchical}.\\

Prior literature partially resolve this problem- however, there are three issues restricting their practicality and detection performance.
\textbf{Issue 1: long-delayed  patterns.}
The malfunctioning effects of abnormal system events often do not manifest immediately.
Kravchik et al. employed LSTM to predict future values based on past values 
and assessed the system status using prediction errors\cite{kravchik2018detecting}.
But, constrained by the capability of LSTM, it is hard to capture long-delayed patterns, which may lead to suboptimal detection performance.
\textit{How can we sufficiently capture such long-delayed patterns?}
\textbf{Issue 2: dynamic changes in sensor-sensor influence.}
Besides long-delayed patterns, the malfunctioning effects may propagate to other sensors.
Wang et al. captured such propagation patterns in water treatment networks by integrating the sensor-sensor connectivity graph for cyber-attack detection~\cite{wang2020defending}.
However, the sensor-sensor influence may shift as the time series changes due to system failures.
Ignoring such dynamics may result in failing to identify propagation patterns and cause poor detection performance.
\textit{How can we consider such dynamic sensor-sensor influence?}
\textbf{Issue 3: high-dimensional data samples.} Considering the labeled data sparsity issue in CPS, existing works focus on unsupervised or semi-supervised setting~\cite{wang2020defending,marti2015anomaly}. 
But traditional models like One-Class SVM are too shallow to fit high-dimensional data samples.
They have substantial time costs for feature engineering and model learning.
\textit{How can we improve the learning efficiency of anomaly detection in high-dimensional scenarios?}\\

To address these, we aim to effectively capture spatial-temporal dynamics in high-dimensional sensor monitoring data.
In CPS, sensors can be viewed as nodes, and their physical connections resemble a graph.
Considering that the monitoring data of each sensor changes over time and that the monitoring data of various sensors influences one another, we model them using a graph stream structure.
Based on that, we propose a new framework called \underline{D}eep \underline{G}raph \underline{S}tream \underline{S}upport \underline{V}ector \underline{D}ata \underline{D}escription (\textbf{DGS-SVDD}).  
Specifically, to capture long-delayed patterns, we first develop a temporal embedding module based on transformer~\cite{DBLP:journals/corr/VaswaniSPUJGKP17}. This module is used to extract these patterns from individual sensor monitoring data and embed them in low-dimensional vectors.
Then, to comprehend dynamic changes in sensor-sensor connection, we estimate the influence between sensors using the previously learned temporal embedding of sensors.
The estimated weight matrix is integrated with the sensor-sensor physically connected graph to produce an enhanced graph. 
We map the temporal embeddings to each node in the enhanced graph as its attributes to form a new attributed graph.
After that, we input this graph into the variational graph auto-encoder (VGAE)~\cite{kipf2016variational} to preserve all information as final spatial-temporal embeddings.
Moreover, to effectively detect anomalies in high-dimensional data, we adopt deep learning to learn the hypersphere that encompasses normal embeddings.
The distances between the hypersphere and data samples are calculated to be criteria to predict the system status at each time segment.
Finally, we conduct extensive experiments on a real-world dataset to validate the superiority of our work. 
In particular, compared to the best baseline model, DGS-SVDD improves F1-score by 35.87\% and AUC by 19.32\%, while accelerating model training and inference by 32 times.

\section{Preliminaries}
\subsection{Definitions}
\begin{definition}
 \textbf{Graph Stream.}
 \iffalse
    \normalfont A graph object can be defined as $\mathbf{T}_i$ = $\{{e}_1,{e}_2,{e}_3\\ \cdots{e}_p \cdots {e}_m\}$ where each item ${e}_p$=  (${S}_j,{S}_k;{i},{w}$) indicates a directed edge from sensor $j$ to sensor $k$, with weight $w$=1 if the sensors are directly connected and 0 otherwise and $S_j$ and $S_k$ stand for the corresponding sensor's monitoring data at timestamp $i$. Graph stream is a collection of graph objects. Hence, we can define the graph stream of length $L_x$ at time segment $t$ as $\mathbf{X}_t = [\mathbf{T}_i, \mathbf{T}_{i+1}, \mathbf{T}_{i+2},\cdots \mathbf{T}_{i+L_{x}-1}]$.
    
\fi
    \normalfont A graph object $\mathcal{G}_i$ describes the monitoring values of the Cyber-Physical System at timestamp $i$. It can be defined as $\mathcal{G}_i$ = ($\mathcal{V}$,$\mathcal{E}$,$\mathbf{t}_i$) where $\mathcal{V}$ is the vertex (i.e., sensor) set with a size of $n$;
    $\mathcal{E}$ is the edge set with a size of $m$, and each edge indicates the physical connectivity between any two sensors; 
    $\mathbf{t}_i$ is a list that contains the monitoring value of $n$ sensors at the $i$-th timestamp.
    A graph stream is a collection of graph objects over the temporal dimension.
    The graph stream with the length of $L_x$ at the $t$-th time segment can be defined as  $\mathbf{X}_t = [\mathcal{G}_i, \mathcal{G}_{i+1}, \cdots \mathcal{G}_{i+L_{x}-1}]$.

    % $V$=$\{0,1,2,\cdots,n\}$, $n$ being the number of sensors in the CPS and $E$ represents the physical connectivity of the sensors. We can define $E$ as $E$ =  $\{{e}_1,{e}_2,{e}_3 \cdots{e}_p \cdots {e}_m\}$ where each item ${e}_p$=  (${S}_j,{S}_k$) indicates a directed edge from sensor $j$ to sensor $k$. The term $a_i$ in $\mathbf{T}_i$ is the collection of sensor monitoring data at timestamp $i$ and for a CPS with $n$ sensors, it can be written as ${Q}_i$ = $\{{s}_0,{s}_1,{s}_2 \cdots{s}_l \cdots {s}_n\}$ where $s_l$ is the reading of sensor $l$ at said timestamp. Graph stream is a collection of graph objects. Hence, we can define the graph stream of length $L_x$ at time segment $t$ as $\mathbf{X}_t = [\mathbf{T}_i, \mathbf{T}_{i+1}, \mathbf{T}_{i+2},\cdots \mathbf{T}_{i+L_{x}-1}]$.
     
    % $\mathcal{W}$ = $\{\mathrm{W}^1,\mathrm{W}^2,... \mathrm{W}^L\}$
\end{definition}

\begin{definition}
 \textbf{Weighted Attributed Graph.}
 \iffalse
    \normalfont The weight term $w$ in each item of a graph object represents the connectivity of two sensors in a CPS. To capture the dynamic changes in sensor-sensor influence, for graph stream at each time segment, we integrate an estimated weight matrix with its objects constructed solely based on the physical connectivity of sensors. A weighted attributed graph generator as discussed in Section \ref{cluster} is utilized to this end, which modifies each item ${e}_p$ in graph object $\mathbf{T}_i$ as ${e}_p$=  (${S}_j,{S}_k;{i},{w_p}$) where ${w}_p$ captures both the physical connectivity and the dynamic changes in influence between sensor $j$ and sensor $k$.
    
\fi
    \normalfont 
    The edge set $\mathcal{E}$ of each graph object in the graph stream $\mathbf{X}_t$ does not change over time, which is a binary edge set that reflects the physical connectivity between sensors.
    However, the correlations between different sensors may change as system failures happen.
    To capture such dynamics, we use $\mathcal{\tilde{G}}_t=(\mathcal{V},\mathcal{\tilde{E}}_t,\mathbf{U}_t)$ to denote the weighted attributed graph at the $t$-th time segment.
    In the graph, $\mathcal{V}$ is the same as the graph object in the graph stream, which is the vertex (i.e., sensor) set with a size of $n$;
    $\mathcal{\Tilde{E}}_t$ is the weighted edge set, in which each item indicates the weighted influence calculated from the temporal information between two sensors;
    $\mathbf{U}_t$ is the attributes of each vertex, which is also the temporal embedding of each node at the current time segment. 
    Thus, $\mathcal{\Tilde{G}}_t$ contains the spatial-temporal information of the system.

    % The term $E$ in graph object $\mathbf{T}_i$ represents the physical connectivity of the sensors in a CPS. To capture the dynamic changes in sensor-sensor influence, for graph stream at each time segment, we utilize a weighted attributed graph generator to modify each term $e_p$ in $E$ as $e_p$ = (${S}_j,{S}_k;{w_p}$) where $w_p$ is the weight of the edge between sensor $j$ and sensor $k$. $w_p$ captures both the connectivity and the dynamic changes in influence between the shared sensors.

\end{definition}

\subsection{Problem Statement}
Our goal is to detect anomalies in cyber-physical systems at each time segment.
Formally, assuming that the graph stream data at the $t$-th segment is $\mathbf{X}_t$, the corresponding system status is $y_t$.
We aim to find an outlier detection function that learns the mapping relation between $\mathbf{X}_t$ and $y_t$, denoted by $f(\mathbf{X}_t) \rightarrow y_t$. Here, $y_t$ is a binary constant whose value is 1 if the system status is abnormal and 0 otherwise.

% is to develop a structured framework to detect anomalies in Cyber-Physical systems. Formally, we model the sensor monitoring data of CPS and their physical connectivity in a graph stream  structure. For each time segment $t$, graph stream $\mathbf{X}_t$ of length $L_x$ is generated such that $\mathbf{X}_t\in\mathbb{R}^{L_{x} \times n}$, where $n$ is the number of sensors in the CPS. Each time segment is associated with a anomaly label; if a graph object $\mathbf{T}_i$ in graph stream $\mathbf{X}_t$ is detected to be anomalous by our framework, the label of said segment is marked as $y_t$=1 and 0 otherwise.

\section{Methodology}
\label{method}
In this section, we give an overview of our framework and then describe each technical part in detail.

% In this section, we’ll start by introducing the time segment embedding of our proposed framework. After that, we’ll illustrate the construction of weighted attributed spatio-temporal graphs(STGs) using the obtained temporal embeddings and sensor networks constructed based on their adjacency and dependency. Then we’ll present the spatio-temporal graph based representation learning and, finally, we’ll discuss the integration of unsupervised one-class detection with Deep SVDD.

\subsection{Framework Overview}
\begin{figure*}[!t]
\vspace{-0.15cm}
    \centering
    \includegraphics[width=1.0\linewidth]{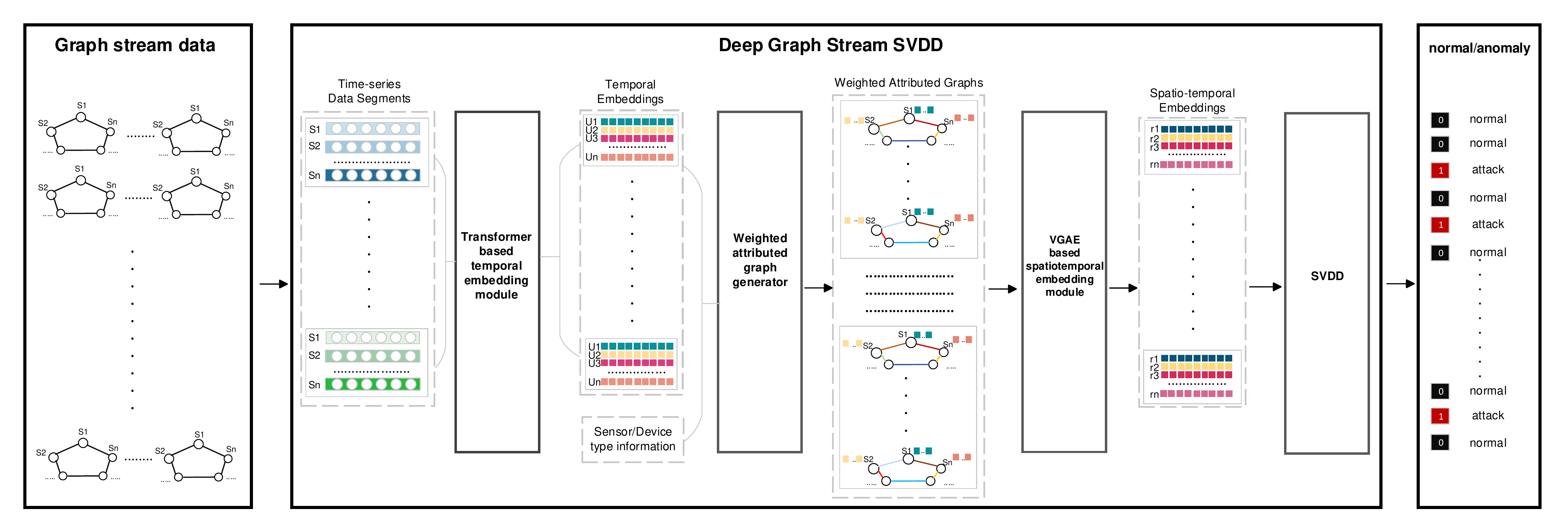}
    \vspace{-0.15cm}
    % \captionsetup{justification=centering}
    \captionsetup{width=0.8\textwidth}
    \vspace{-0.43cm}
    \caption{{
    An overview of our framework. There are four key components: transformer-based temporal embedding module, weighted attributed graph generator, VGAE-based spatiotemporal embedding module, and SVDD-based outlier detector.
    }  
    }
    \vspace{-0.5cm}
    \label{fig:framework}
\end{figure*}

Figure~\ref{fig:framework} shows an overview of our framework, named DGS-SVDD.
Specifically, we start by feeding the DGS-SVDD model the graph stream data for one time segment.
In the model, we first analyze the graph stream data by  
adopting the transformer-based temporal embedding module to extract temporal dependencies.
Then, we use the learnt temporal embedding to estimate the dynamics of sensor-sensor influence and combine it with information about the topological structure of the graph stream data to generate weighted attributed graphs.
We then input the graph into the variational graph autoencoder (VGAE)-based spatial embedding module to get the spatial-temporal embeddings.
Finally, we estimate the boundary of the embeddings of normal data using deep learning and support vector data description (SVDD), and predict the system status by measuring how far away the embedding sample is from the boundary.

\subsection{Embedding temporal patterns of the graph stream data}
\label{temporal}

The temporal patterns of sensors may evolve over time if abnormal system events occur.
We create a temporal embedding module that uses a transformer in a predictive manner to capture such patterns for accurate anomaly detection.
To illustrate the following calculation process, we use the graph stream data $\mathbf{X}_t$ at the $t$-th time segment as an example.
We ignore the topological structure of the graph stream data at first during the temporal embedding learning process.
Thus, we collect the time series data in $\mathbf{X}_t$ to form a temporal matrix $\mathbf{T}_t=[\mathbf{t}_1,\mathbf{t}_2,\cdots,\mathbf{t}_{L_x}]$, such that $\mathbf{T}_t\in\mathbb{R}^{n \times L_{x}}$, where $n$ is the number of sensors and $L_x$ is the length of the time segment.

The temporal embedding module consists of an encoder and a decoder.
For the encoder part, 
we input $\mathbf{T}_t$ into it for learning enhanced temporal embedding $\mathbf{U}_t$.
Specifically, we first use the multi-head attention mechanism to calculate the attention matrices between $\mathbf{T}_t$  and itself for enhancing the temporal patterns among different sensors by information sharing.
Considering that the calculation process in each head is the same, we take \textit{head$_1$} as an example to illustrate.
To obtain the self-attention matrix $\text{Attn}(\mathbf{T}_t,\mathbf{T}_t)$, we input $\mathbf{T}_t$ into \textit{head$_1$}, which can be  formulated as follows,
\begin{equation}
       \text{Attn}(\mathbf{T}_t,\mathbf{T}_t) = \textit{softmax}(\frac{(\mathbf{T}_t\cdot\mathbf{W}^Q_t)(\mathbf{T}_t\cdot\mathbf{W}^K_t)^\top}{\sqrt{L_x}})\cdot (\mathbf{T}_t\cdot\mathbf{W}^{V}_{t}) 
\end{equation}
where 
$\mathbf{W}^{K}_{t}\in\mathbb{R}^{L_x\times d}$, $\mathbf{W}^{Q}_{t}\in\mathbb{R}^{L_x\times d}$, and $\mathbf{W}^{V}_{t}\in\mathbb{R}^{L_x\times d}$ are the weight matrix for ``key'', ``query'' and ``value'' embeddings; ${\sqrt{L_x}}$ is the scaling factor.
Assuming that we have $h$ heads, we concatenate the learned attention matrix together in order to capture the temporal patterns of monitoring data from different perspectives.
The calculation process can be defined as follows:
\begin{equation}
    \mathbf{T}'_{t} = \text{Concat} (\text{Attn}^1_t, \text{Attn}^2_t,\cdots, \text{Attn}^h_t) \cdot \mathbf{W}^O_t
\end{equation}
where $\mathbf{W}^O_t\in\mathbb{R}^{hd\times d_\text{model}}$ is the weight matrix and $\mathbf{T}'_{t}\in\mathbb{R}^{n\times d_\text{model}}$. 
After that, we input $\mathbf{T}'_{t}$ into a fully connected feed-forward network constructed by two linear layers to obtain the enhanced embedding $\mathbf{U}_t \in \mathbb{R}^{n\times d_\text{model}}$. 
The calculation process can be defined as follows:
\begin{equation}
    \mathbf{U}_{t} = \mathbf{T}'_t + \text{Relu}(\mathbf{T}'_t \cdot \mathbf{W}_t^1 + \mathbf{b}_t^1)\cdot \mathbf{W}_t^2 + \mathbf{b}_t^2
\end{equation}
where $\mathbf{W}_t^1$ and $\mathbf{W}_t^2$ are the weight matrix respectively and their shape information is $\mathbb{R}^{d_\text{model}\times d_\text{model}}$;
$\mathbf{b}^1_t$ and $\mathbf{b}^2_t$ are the bias item respectively and their shape information is $\mathbb{R}^{n\times d_\text{model}}$.

For the decoder part, we input the learned embedding $\mathbf{U}_t$ into a prediction layer to predict the monitoring value of the future time segment.
The prediction process can be defined as follows:
\begin{equation}
    \mathbf{\Check{T}}_{t+1} = \mathbf{U}_t\cdot\mathbf{W}_t^p + \mathbf{b}_t^p
\end{equation}
where $\mathbf{\check{T}}_{t+1}\in\mathbb{R}^{n\times L_x}$ is the prediction value of the next time segment; $\mathbf{W}_t^p\in\mathbb{R}^{d_\text{model}\times L_x}$ is the weight matrix and $\mathbf{b}_t^p \in \mathbb{R}^{n\times L_x}$ is the bias item.
During the optimization process, we minimize the difference between the prediction $\mathbf{\Check{T}}_{t+1}$ and the real monitoring value $\mathbf{T}_{t+1}$.
The optimization objective can be defined as
follows
\begin{equation}
    \text{min}\sum_{t=1}^{L_x} ||\mathbf{T}_{t+1}-\mathbf{\check{T}}_{t+1}||^{2}
\end{equation}
When the model converges, we have preserved temporal patterns of monitoring data in the temporal embedding $\mathbf{U}_t$.

\subsection{Generating dynamic weighted attributed graphs}
\label{cluster}
In CPS, different sensors connect with each other, which forms a sensor-sensor graph.
As a result, the malfunctioning effects of system abnormal events may propagate over time following the graph structure.
But, the sensor-sensor influence is not static and may vary as the monitoring data changes are caused by system anomaly events.
To capture such dynamics, we want to build up weighted attributed graphs using sensor-type information and learned temporal embeddings.
For simplicity, we take the graph stream data of $t$-th time segment $\mathbf{X}_t$ as an example to illustrate the following calculation process.

Specifically, the adjacency matrix of $\mathbf{X}_t$  is $\mathbf{A} \in \mathbb{R}^{n\times n}$, which reflects the physical connectivity between different sensors. 
$\mathbf{A}[i,j]$ = 1 when sensor $i$ and $j$ are directly connected and $\mathbf{A}[i,j]$ = 0 otherwise.
From section \ref{temporal}, we have obtained the temporal embedding $\mathbf{U}_t \in\mathbb{R}^{n \times d_{model}}$,  each row of which represents the temporal embedding for each sensor.
We assume that the sensors belonging to the same type have similar changing patterns when confronted with system anomaly events. 
Thus, we want to capture this characteristic by integrating sensor type information into the adjacency matrix.
We calculate the sensor type embedding by averaging the temporal embedding of sensors belonging to the type.
After that, we construct a type-type similarity matrix $\mathbf{C}_t\in \mathbb{R}^{k\times k}$ by calculating the 
cosine similarity between each pair of sensor types, $k$ being the number of sensor types.
Moreover, we construct the similarity matrix $\mathbf{\check{C}}_t\in \mathbb{R}^{n \times n}$ by mapping $\mathbf{C}_t$ to each element position of $\mathbf{A}$. 
For instance, if sensor 1 belongs to type 2 and sensor 2 belongs to type 3, we update $\mathbf{\check{C}}_t[1,2]$ with $\mathbf{C}_t[2,3]$.
We then introduce the dynamic property to the adjacency matrix $\mathbf{A}$ through element-wise multiplication between $\mathbf{A}$ and $\mathbf{\Check{C}}_t$.
Each temporal embedding of this time segment is mapped to the weighted graph as the node attributes according to sensor information.
The obtained weighted attributed graph $\mathcal{\tilde{G}}_t$ contains all spatial-temporal information of CPS for the $t$-th time segment. 
The topological influence of this graph may change over time.

\subsection{Representation learning for weighted attributed graph}
\label{spatial}
To make the outlier detection model easily comprehend the information of $\mathcal{G}_t$,  we develop a representation learning module based on variational graph autoencoder (VGAE).
For simplicity, we use $\mathcal{G}_t$ to illustrate the representation learning process.
For $\mathcal{G}_t=(\mathcal{V},\mathcal{\tilde{E}}_t,\mathbf{U}_t)$ ,
the adjacency matrix  is $\mathbf{\tilde{A}}_t$ made up by $\mathcal{V}$ and $\mathcal{\tilde{E}}_t$, and the feature matrix is $\mathbf{U}_t$.

Specifically, this module follows the encoder-decoder paradigm. 
The encoder includes two Graph Convolutional Network(GCN) layers. 
The first GCN layer takes $\mathbf{U}_t$ and $\mathbf{\tilde{A}}_t$ as inputs and outputs a lower dimensional feature matrix $\mathbf{\hat{U}}_t$.  
The calculation process can be represented as follows:
\begin{equation}
\begin{aligned}
    \mathbf{\hat{U}}_t  & = \text{Relu}({\mathbf{\hat{D}}_t^{-1/2}}\mathbf{\tilde{A}}_t{\mathbf{\hat{D}}_t^{-1/2}}\mathbf{U}_t\mathbf{\tilde{W}}_0)
\end{aligned}
\end{equation}
where $\mathbf{\hat{D}}_t$ is the diagonal degree matrix of $\mathcal{G_t}$ and $\mathbf{\tilde{W}}_0$ is the weight matrix of the first GCN layer. 
The second GCN layer estimates the distribution of the graph embeddings.
Assuming that such embeddings conform to the normal distribution $\mathcal{N}(\bm{\mu}_t,\bm{\delta}_t)$, we  need to estimate the mean $\bm{\mu}_t$ and variance $\bm{\delta}_t$ of the distribution.
Thus, the encoding process of the second GCN layer can be formulated as follows:
\begin{equation}
\begin{aligned}
    \bm{\mu}_t, \textit{log}(\bm{\delta}_t^2) 
                       & = \text{Relu}({\mathbf{\hat{D}}_t^{-1/2}}\mathbf{A}_t{\mathbf{\hat{D}}_t^{-1/2}}\mathbf{\hat{U}}_t\mathbf{\Tilde{W}}_1)
\end{aligned}
\end{equation}
where $\mathbf{\tilde{W}}_1$ is the weight matrix of the second GCN layer.
Then, we use the reparameterization technique to mimic the sample operation to 
obtain the graph embedding $\mathbf{r}_t$, which can be represented as follows:
\begin{equation}
    \mathbf{r}_t = \bm{\mu}_t + \bm{\delta}_t \times \bm{\epsilon}_t
\end{equation}
where $\bm{\epsilon}_t$ is the random variable vector, which is sampled from $ \mathcal{N}(0,I)$. Here, $\mathcal{N}(0,I)$ represents the high-dimensional standard normal distribution. 

The decoder part aims to reconstruct the adjacency matrix of the graph using $\mathbf{r}_t$, which can be defined as follows:
\begin{equation}
    \mathbf{\hat{A}}_t = \sigma(\mathbf{r}_t{\mathbf{r}_t}^\top)
\end{equation}
where $\mathbf{\hat{A}}_t$ is the reconstructed adjacency matrix and $\mathbf{r}_t{\mathbf{r}_t}^\top$ = $||\mathbf{r}_t||$ $||{\mathbf{r}_t}^\top||$cos $\theta$.

During the optimization process, we aim to minimize two objectives: 1) the divergence between the prior embedding distribution $\mathcal{N}(0,I)$ and the estimated embedding distribution $\mathcal{N}(\bm{\mu}_t,\bm{\delta}_t)$; 2) the difference between the adjacency matrix $\mathbf{A}_t$ and the reconstructed adjacency matrix $\mathbf{\tilde{A}}_t$;
Thus, the optimization objective function is as follows:
\begin{equation}
\text{min}\sum_{t=1}^{T}\underbrace{\textit{KL}[q(\mathbf{r}_t|\mathbf{U}_t, \mathbf{A}_t)||p(\mathbf{r}_t)]}_\text{KL divergance between $q(.)$ and $p(.)$} + \overbrace{||\mathbf{A_{t}}-\mathbf{\hat{A}_{t}}||^{2}}^\text{Loss between $\mathbf{A}_t$ and $\mathbf{\hat{A}}_t$}
\label{loss_g}    
\end{equation}
where \textit{KL} refers to the Kullback-Leibler divergence; $q(.|.)$ is the estimated embedding distribution and $p(.)$ is the prior embedding distribution.
When the model converges, the graph embedding $\mathbf{r}_t \in \mathbb{R}^{n\times d_{\text{emb}}}$ contains spatiotemporal patterns of the monitoring data for the $t$-th time segment.

\subsection{One-Class Detection with SVDD}
\label{DeepSVDD}

Considering the sparsity issue of labeled anomaly data in CPS, anomaly detection is done in an unsupervised setting.
Inspired by deep SVDD~\cite{ruff2018deep},  we aim to learn a hypersphere that encircles most of the normal data, with data samples located beyond it being anomalous.
Due to the complex nonlinear relations among the monitoring data, we use deep neural networks to approximate this hypersphere.

Specifically, through the above procedure, we collecte the spatiotemporal embedding of all time segments, denoted by $\left[\mathbf{r}_1, \mathbf{r}_2, \cdots, \mathbf{r}_T \right]$.
 We input them into multi-layer neural networks to estimate the non-linear hypersphere. 
Our goal is to minimize the volume of this data-enclosing hypersphere.
The optimization objective can be defined as follows:
\begin{equation}
    \label{svdd_objective}
    \min_{\mathcal{W}}\underbrace{\frac{1}{n} \sum_{t=1}^{T}||\phi(\mathbf{r}_t ; \mathcal{W})-c||^{2}}_{\substack{\text{Average sum of weights, using} \\ \text{squared error, for all normal} \\ \text {training instances (from T segments)}}} + \overbrace{\frac{\lambda}{2} ||\mathcal{W}||_F^2}^\text{Regularization item}
\end{equation}
where $\mathcal{W}$ is the set of weight matrix of each neural network layer;
$\phi(\mathbf{r}_t ; \mathcal{W})$ maps $\mathbf{r}_t$ to the non-linear hidden representation space;
$c$ is the predefined hypersphere center;
$\lambda$ is the weight decay regularizer. 
The first term of the equation aims to find the most suitable hypersphere that has the closest distance to the center $c$.
The second term is to reduce the complexity of $\mathcal{W}$, which avoids overfitting.
As the model converges, we get the network parameter for a trained model, $\mathcal{W}^*$.

During the testing stage, given the embedding of a test sample $\mathbf{r}_o$, we input it into the well-trained neural networks to get the new representation. 
Then, we calculate the anomaly score of the sample based on the distance between it and the center of the hypersphere.
The process can be formulated as follows:
\begin{equation}
\label{anam_score}
    s(\mathbf{r}_o) = ||\phi(\mathbf{r}_o ; \mathcal{W}^*)-c||^{2}
\end{equation}
After that, we compare the score with our predefined threshold to assess the abnormal status of each time segment in CPS.

\section{Experiments}
We conduct extensive experiments to validate the efficacy and efficiency of our framework (DGS-SVDD) and the necessity of each technical component.

\iffalse
We conduct experiment in an attempt to answer the following research questions:
1. Can we utilize an anomaly detector such as Deep SVDD that has been specially crafted towards computer vision, to monitor abnormalities in crucial cyber-physical systems from graph streams of data?\\
2. Does the proposed dynamic spatio-temporal representation learning component of our framework enhance the anomaly detection performance?\\
3. Is our proposed model robust to different size of training data and is it sensitive to changes in parameters discussed in Section \ref{method}?\\
4. How much time does our method cost and how fast is it compared to traditional anomaly detection methods?
\fi

\subsection{Experimental Settings}
\subsubsection{Data Description}
\label{swat_details}
We adopt the SWaT dataset~\cite{mathur2016swat}, from the Singapore University of Technology and Design in our experiments.
This dataset was collected from a water treatment testbed that contains 51 sensors and actuators.
The collection process continued for 11 days.
The system's status was normal for the first 7 days and for the final 4 days, it was attacked by a cyber-attack model.
The statistical information of the SWaT dataset is shown in Table~\ref{swat_stat}.
Our goal is to detect attack anomalies as precisely as feasible. We only use the normal data to train our model.
After the training phase, we validate the capability of  our model by detecting the status of the testing data that contains both normal and anomalous data.

\iffalse
To evaluate our proposed method, we use the secure water treatment system (SWaT) dataset\cite{mathur2016swat} from the Singapore University of Technology and Design. Initially launched in May 2015, The dataset is collected from a water treatment testbed for cyber-attack investigation. The SWaT dataset collection process lasted for 11 days with the system operating for 24 hours each day, such that the network traffic and the values obtained from all 51 sensors and actuators are recorded. Table \ref{swat_stat} summarizes the statistics of this dataset that includes a “normal” set comprising of information gathered from the water treatment’s normal state. The “attack” set on the other hand, consists of information regarding cyber-attacks that were launched during the last 4 days of the collection process. These attacks were launched with different intents and diverse lasting duration (from a few minutes to an hour). 
\fi

\begin{table}[htbp!]
\vspace{-0.7cm}
\fontsize{8.5pt}{8.5pt}
\selectfont
\centering
\caption{Statistics of SWaT Dataset}
\label{swat_stat}
% \resizebox{\textwidth}{!}
{%
\begin{tabular}{@{}cccccc@{}}
\toprule
Data Type  & \ Feature Number & Total Items & \ Anomaly Number & Normal/Anomaly \\ \midrule
Normal    & 51             & 496800      & 0               & -              \\
Anomalous  & 51             & 449919      & 53900           & 7:1            \\ \bottomrule
\end{tabular}%
}
\vspace{-1.2cm}
\end{table}

\subsubsection{Evaluation Metrics}
\label{eval_met}
We evaluate the model performance in terms of precision, recall, area under the receiver operating characteristic curve (ROC/AUC), and F1-score. 
We adopt the point-adjust way to calculate these metrics. 
In particular,
abnormal observations typically occur in succession to generate anomaly segments and an anomaly alert can be triggered inside any subset of a real window for anomalies.
Therefore, if one of the observations in an actual anomaly segment is detected as abnormal, we would consider the time points of the entire segment to have been accurately detected.

\subsubsection{Baseline Models}
To make the comparison objective, we input the spatial-temporal embedding vector $\mathbf{r}_t$
into baseline models instead of the original data.
There are seven baselines in our work:
\textbf{KNN}~\cite{peterson2009k}: calculates the anomaly score of each sample according to the anomaly situation of its K nearest neighborhoods.
\textbf{Isolation-Forest}\cite{liu2008isolation}: estimates the average path length (anomaly score) from the root node to the terminating node for isolating a data sample using a collection of trees.\textbf{LODA}\cite{pevny2016loda}: collects a list of weak anomaly detectors to produce a stronger one. LODA can process sequential data flow and is robust to missing data. 
\textbf{LOF}\cite{breunig2000lof}:
measures the anomalous status of each sample based on its local density. 
If the density is low, the sample is abnormal; otherwise, it is normal.
\textbf{ABOD}\cite{kriegel2008angle}: is an angle-based outlier detector. If a data sample is located in the same direction of more than 
 K data samples, it is an outlier; otherwise it is normal data.
 \textbf{OC-SVM}\cite{manevitz2001one}: finds a hyperplane to divide normal and abnormal data through kernel functions..
 \textbf{GANomaly}\cite{akcay2018ganomaly}: utilizes an  encoder-decoder-encoder architecture. 
It evaluates the anomaly status of each sample by calculating the difference between the output embedding of two encoders.
\begin{table}[]
\centering
\fontsize{9.8pt}{9.3pt}
\selectfont
% \scriptsize
\caption{Experimental Results on SWaT dataset}
\label{performance}
 % \resizebox{\textwidth}{!}
 {%
\begin{tabular}{cccccc}
\hline
Method             & Precision (\%) & Recall (\%)    & F1-score (\%)  & AUC (\%)       \\ \hline
OC-SVM                     & 34.11          & 68.23          & 45.48          & 75             \\
Isolation-Forest           & 35.42          & 81.67          & 49.42          & 80             \\
LOF                        & 15.81          & 93.88          & 27.06          & 63             \\
KNN                        & 15.24          & 96.77          & 26.37          & 61             \\
ABOD                       & 14.2           & \textbf{97.93} & 24.81          & 58             \\
GANomaly                   & 42.12          & 67.87          & 51.98          & 68.64          \\
LODA                       & 75.25          & 38.13          & 50.61          & 67.1           \\
DGS-SVDD       & \textbf{94.17} & 82.33          & \textbf{87.85} & \textbf{87.96} \\ \hline
\end{tabular}%
}
\vspace{-0.6cm}
\end{table}

 \subsection{Experimental Results}
 \subsubsection{Overall Performance}

Table \ref{performance} shows experimental results on the SWaT dataset, with the best scores highlighted in \textbf{bold}. As can be seen, DGS-SVDD outperforms other baseline models in the majority of evaluation metrics. Compared with the second-best baseline, DGS-SVDD improves precision by 19\%, F1-score by 36\% and AUC by 8\%. This observation validates that DGS-SVDD is effective to detect anomalies accurately. The underlying driver for the success of our model is that DGS-SVDD can capture long-delayed temporal patterns and dynamic sensor-sensor influences in CPS. Another interesting observation is that the detection performance of distance-based or angle-based outlier detectors is poor. A possible reason is that these geometrical measurements are vulnerable to high-dimensional data samples. 

\subsubsection{Ablation Study}
% \usepackage{multirow}
% \usepackage{graphicx}
% The goal of this section is to answer the following question: \textit{Does the proposed dynamic spatio-temporal representation learning component of our framework enhance the anomaly detection performance?} 
To study the individual contribution of each component of DGS-SVDD, we perform ablation studies, the findings of which are summarized in Table \ref{ablation_table} where \textbf{bold} indicates the best score. 
We build four variations of the DGS-SVDD model: 1) We feed unprocessed raw data into SVDD; 2) We only capture temporal patterns; 3) We capture the dynamics of sensor-sensor impact and spatial patterns in CPS; 4) We capture spatial-temporal patterns in CPS but discard the dynamics of sensor-sensor influence.
We can find that DGS-SVDD outperforms its variants by a significant margin.
The observation validates that each technical component of our work is indispensable.
Another interesting observation is that removing the temporal embedding module dramatically degrades the detection performance, rendering the temporal embedding module the highest significance. 
Results from the final experiment show that capturing the dynamics of sensor-sensor influence really boosts model performance.
\begin{table}[]
\vspace{-0.5cm}
\centering
\caption{Ablation Study of DGS-SVDD}
\label{ablation_table}
\resizebox{\textwidth}{!}{%
\begin{tabular}{ccc|cccc}
\hline
\multicolumn{3}{c|}{Method} & \multicolumn{1}{l}{\multirow{2}{*}{Precision (\%)}} & \multicolumn{1}{l}{\multirow{2}{*}{Recall (\%)}} & \multicolumn{1}{l}{\multirow{2}{*}{F1-score (\%)}} & \multicolumn{1}{l}{\multirow{2}{*}{AUC (\%)}} \\ \cline{1-3}
\multicolumn{1}{c|}{\begin{tabular}[c]{@{}c@{}}Transformer-based Temporal\\ Embedding Module\end{tabular}} & \multicolumn{1}{c|}{\begin{tabular}[c]{@{}c@{}}Weighted Attributed\\ Graph Generator\end{tabular}} & \begin{tabular}[c]{@{}c@{}}VGAE-based Spatiotemporal\\ Embedding Module\end{tabular} & \multicolumn{1}{l}{} & \multicolumn{1}{l}{} & \multicolumn{1}{l}{} & \multicolumn{1}{l}{} \\ \hline
\xmark & \xmark & \xmark & 4.61 & 12.45 & 6.74 & 18.55 \\
\cmark & \xmark & \xmark & 69.98 & 64.75 & 67.26 & 78.14 \\
\xmark & \cmark & \cmark & 12.16 & \textbf{99.99} & 21.68 & 18.22 \\
\cmark & \xmark & \cmark & 87.79 & 76.68 & 81.86 & 82.45 \\
\cmark & \cmark & \cmark & \textbf{94.17} & 82.33 & \textbf{87.75} & \textbf{87.96} \\ \hline
\end{tabular}%
}
\vspace{-1cm}
\end{table}

\subsubsection{Robustness Check and Parameter Sensitivity}
\begin{figure}[]
\begin{center}
\subcapraggedrighttrue
\subcaphangtrue
    \vspace{-0.4cm}
    \subfigure[{Varying size of training data}]{\includegraphics[width=3.6cm]{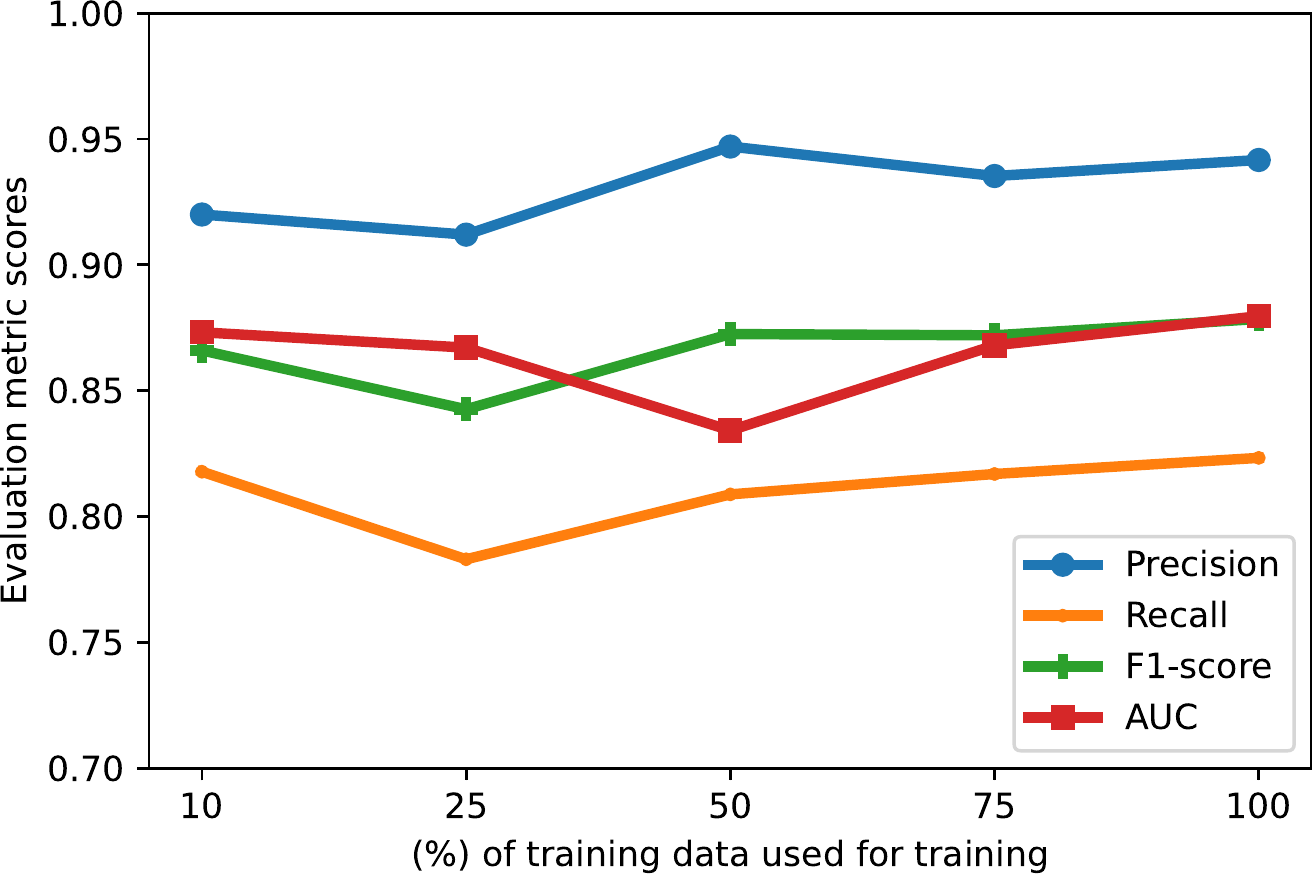} \label{fig:percent}}\hspace{3mm}
    \subfigure[{Varying length of sliding time window}]{\includegraphics[width=3.6cm]{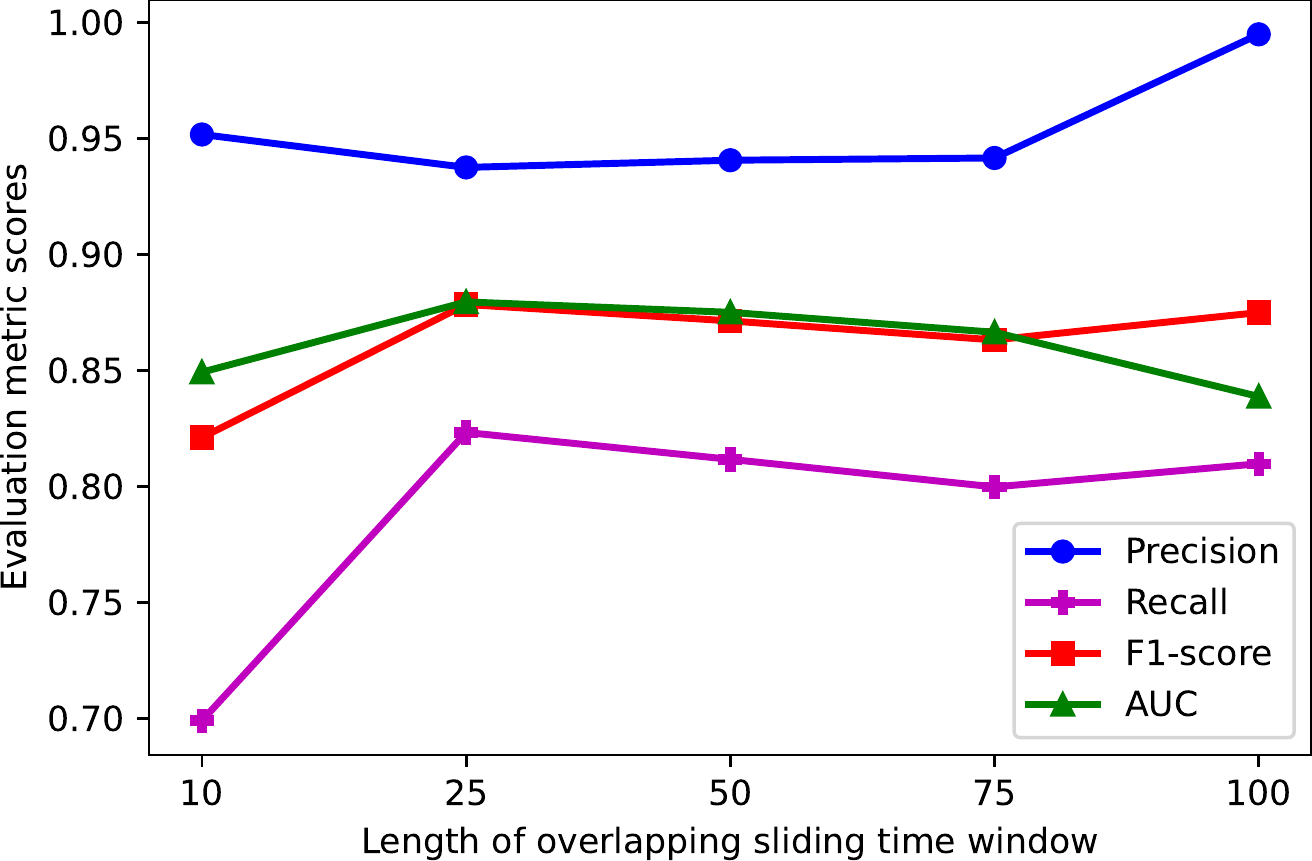}\label{fig:window_size}}\hspace{3mm}
    \subfigure[{Varying length of final embedding}]{\includegraphics[width=3.6cm]{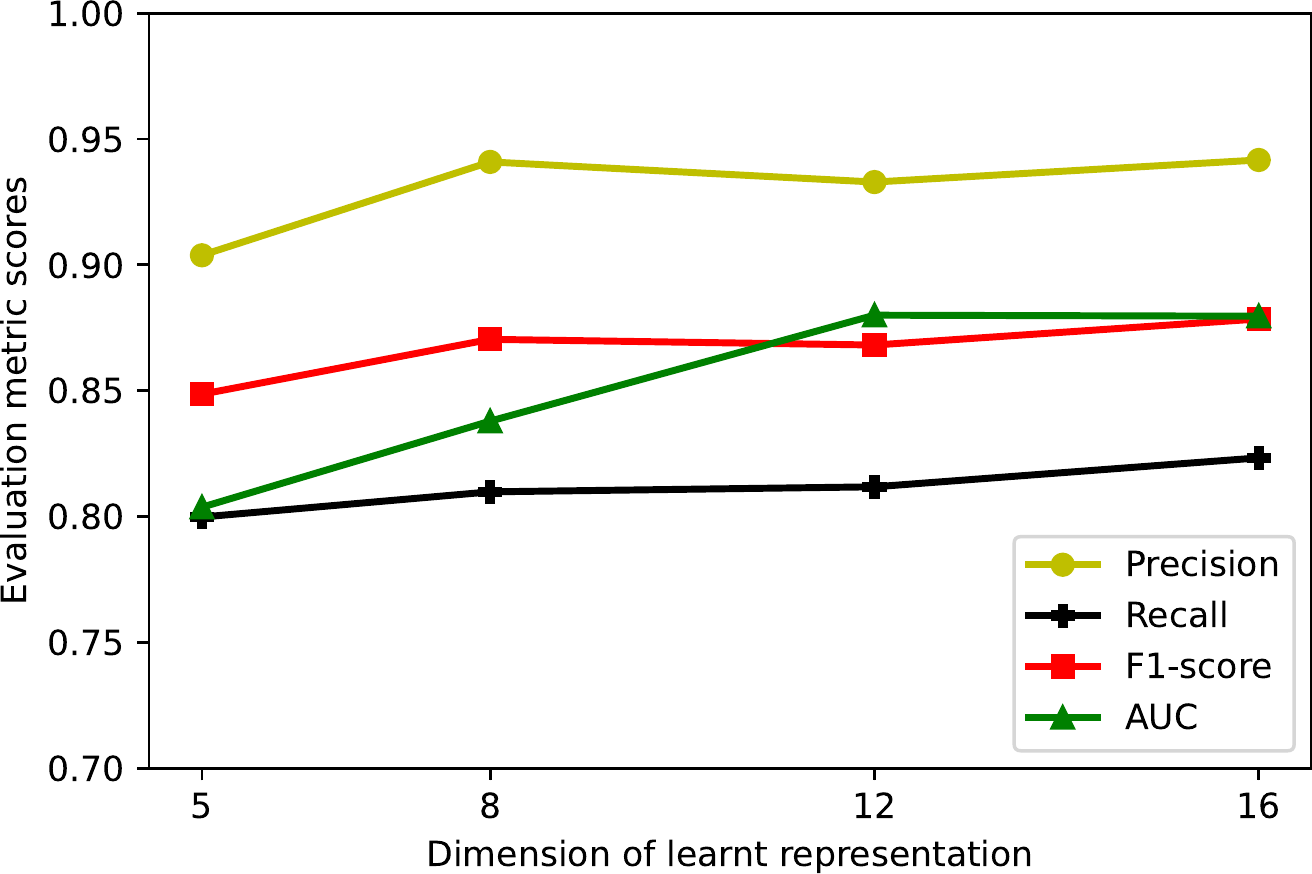}\label{fig:dim_val}}
    \vspace{-0.4cm}
    \caption{Experimental results for robustness check and parameter sensitivity}
    \label{fig:robustness}
    \end{center}
    \vspace{-0.5cm}
\end{figure}

Figure~\ref{fig:robustness} shows the experimental results for robustness check and parameter sensitivity analysis.
To check the model's robustness, we train DGS-SVDD on different percentages of the training data, starting from 10\% to 100\%.
We can find that DGS-SVDD is stable when confronted with different training data from Figure~\ref{fig:percent}.
But, compared with other percentages, DGS-SVDD achieves the best performance when we train it on 50\% training data.
In addition, we vary the dimension of the final spatial-temporal embedding in order to check its impacts.
From Figure~\ref{fig:window_size} and \ref{fig:dim_val}, we can find that DGS-SVDD is barely sensitive to the the sliding window length and dimension of the spatiotemporal embeddings.
This observation validates that DGS-SVDD is robust to the dimension parameters.
A possible reason is that our representation learning module has sufficiently captured spatial-temporal patterns of monitoring data for anomaly detection.

\begin{figure}[]
    \vspace{-0.2cm}
    \centering
    \subfigure[Training time cost]{\includegraphics[width=5.4cm]{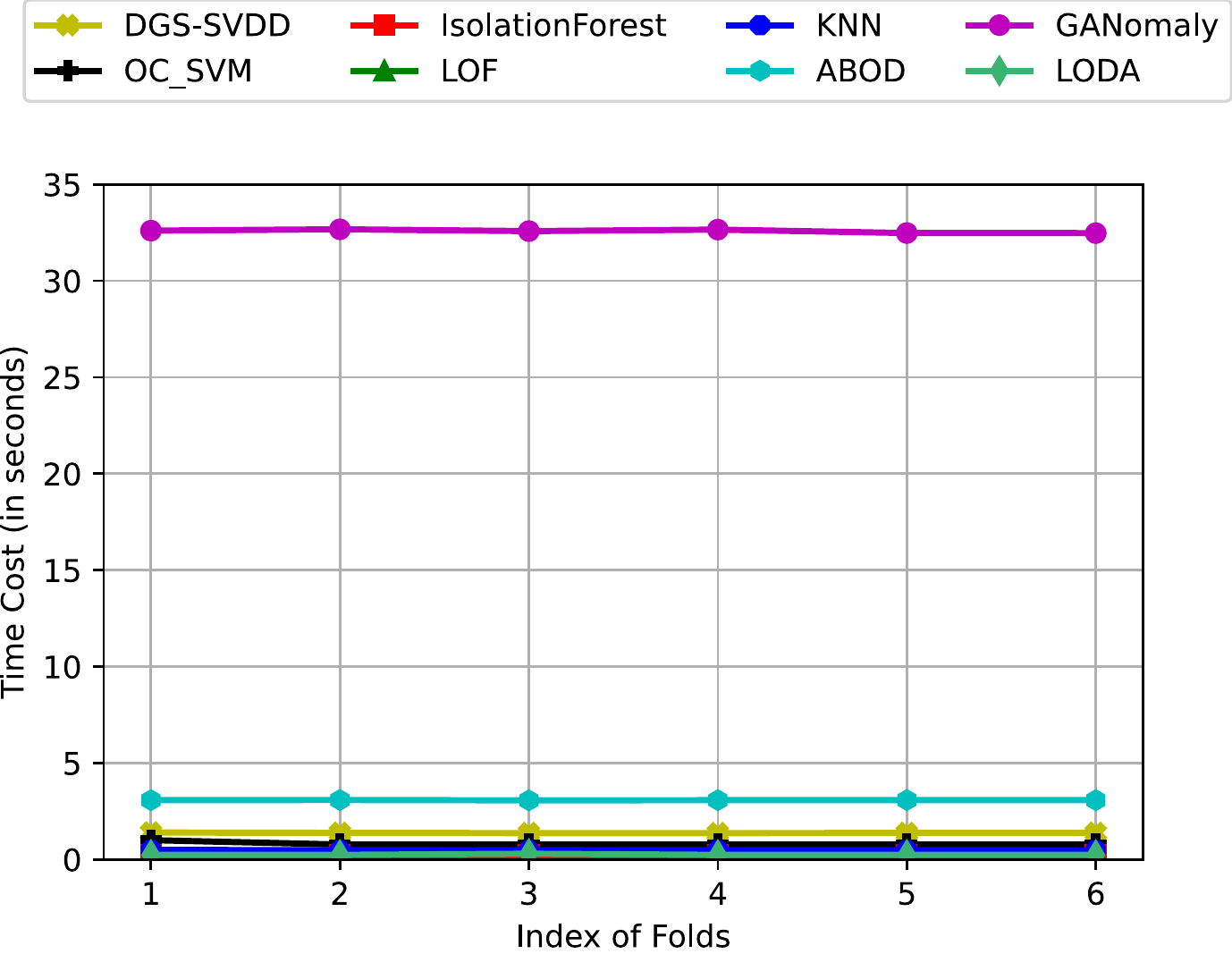} \label{fig:train}}\hspace{3mm}
    \subfigure[Testing time cost]{\includegraphics[width=5.4cm]{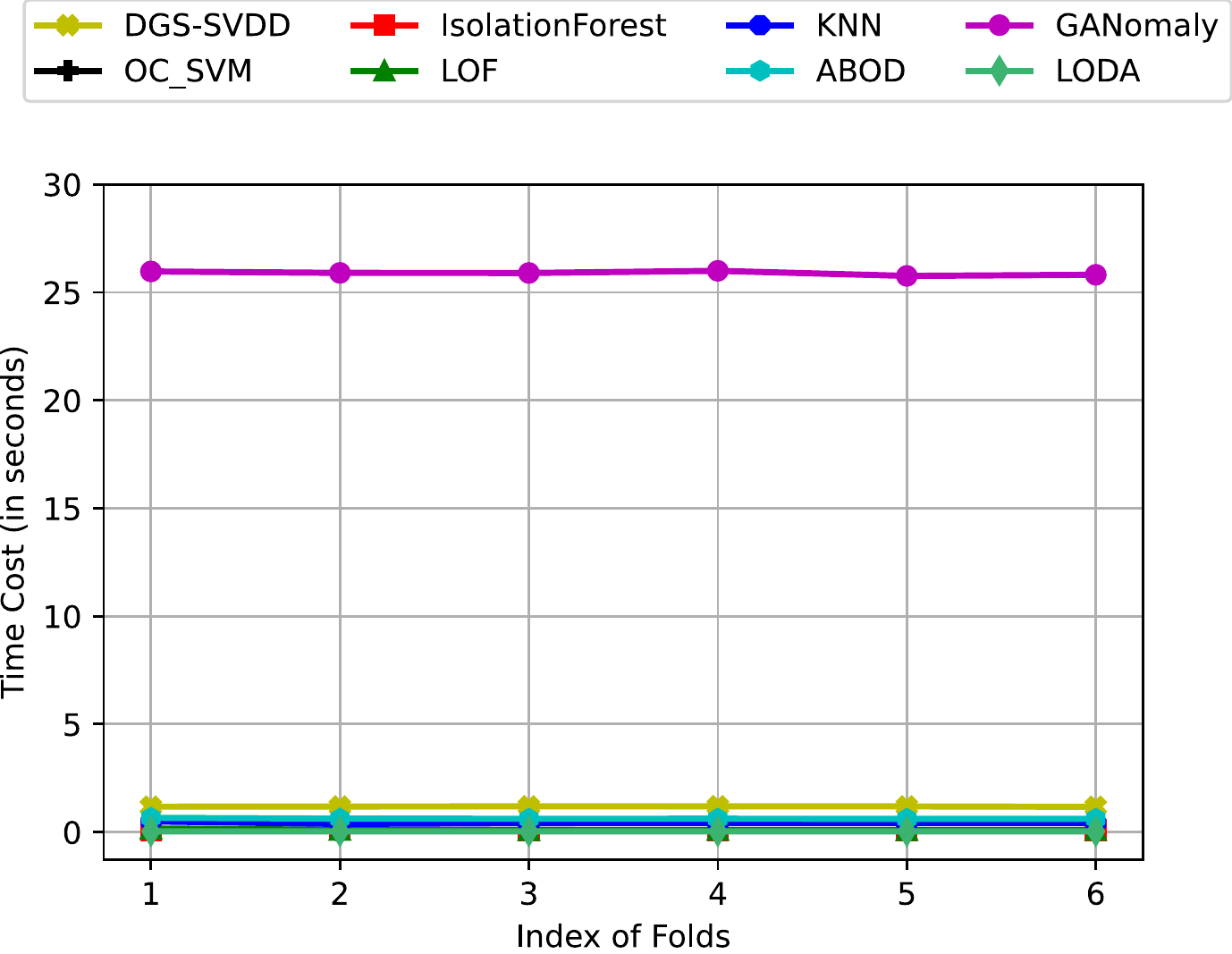}\label{fig:test}}\hspace{3mm}
    % \subfigure[]{\includesvg[width=3.8cm]{dim_val.svg}\label{fig:dim_val}}
    \vspace{-0.4cm}
    \caption{Comparison of different models in terms of training and testing time cost}
    \label{fig:time}
    \vspace{-0.5cm}
\end{figure}
\vspace{-0.3cm}
\subsubsection{Study of Time Cost}
We conduct six folds cross-validation to evaluate the time costs of different models.
Figure \ref{fig:time} illustrates
the comparison results.
We can find that DGS-SVDD can be trained at a time competitive with simple models like OC-SVM or LOF while outperforming them by a huge margin as seen from Table \ref{performance}. 
This shows that DGS-SVDD effectively learns the representation of each time segment of the graph stream data.
Another important observation is that the testing time of DGS-SVDD is consistent with the simpler baselines. 
A potential reason is that the network parameter $\mathcal{W^*}$, as discussed in section \ref{DeepSVDD}, completely characterizes our one-class classifier. This allows fast testing by simply evaluating the network $\phi$ with learnt parameters $\mathcal{W^*}$.

\vspace{-0.3cm}
\section{Related Work}
\textbf{\textit{Anomaly Detection in Cyber-Physical Systems}}. Numerous existing literature have studied the exploitation of temporal and spatial relationships in data streams from CPS to detect anomalous points \cite{kravchik2018detecting}.
For instance, \cite{kravchik2018detecting,li2019mad} adopts a convolutional layer as the first layer of a Convolutional Neural Network to obtain correlations of multiple sensors in a sliding time window. Further, the extracted features are fed to subsequent layers to generate output scores. 
% Meanwhile \cite{canizo2019multi} and \cite{wu2018weighted} utilized Recurrent Neural Network to take the output of the CNN layer and form the prediction layer. 
\cite{li2019mad} proposed a GAN-based framework to capture the spatial-temporal correlation in multidimensional data. 
Both generator and discriminator are utilized to detect anomalies by reconstruction and discrimination errors. 

% Many existing literature have studied the exploitation of temporal and spatial relationships in data streams from CPS to detect anomalous points. Convolutional Neural Networks can extract features of multi-dimensional data jointly via convolution operations. Several approaches\cite{canizo2019multi} \cite{kravchik2018detecting} \cite{wu2018weighted} adopt a convolutional layer as the first layer of the neural network to obtain correlations of multiple sensors in a sliding time window. Further, the extracted features are fed to subsequent layers to generate output scores. Meanwhile \cite{canizo2019multi} and \cite{wu2018weighted} utilized Recurrent Neural Network to take the output of the CNN layer and form the prediction layer. \cite{li2019mad} proposed a GAN-based framework to capture the spatial-temporal correlation in the multidimensional data. Both generator and discriminator are utilized to detect anomalies by reconstruction and discrimination errors. LSTM models are used to build the generator and discriminator. The framework takes the graph stream as input and aims to detect false control signals.

\textbf{\textit{Outlier detection with Deep SVDD}}. After being introduced in \cite{ruff2018deep}, deep SVDD and its many variants have been used for deep outlier detection. \cite{zhang2021anomaly} designed \textit{deep structure preservation SVDD} by integrating deep feature extraction with the data structure preservation. \cite{zhou2021vae} proposed a \textit{Deep SVDD-VAE}, where VAE is used to reconstruct the input sequences while a spherical discriminative boundary is learned with the latent representations simultaneously, based on SVDD. 
% \cite{zhang2021towards} introduced \textit{Deep Fair SVDD} where the model is trained by using an adversarial network to de-correlate the relationships between the sensory attributes and the learned representations.
Although these models have been successfully applied to detect anomalies in the domain of computer vision, this domain lacks temporal and spatial dependencies prevalent in graph stream data generated from CPS.

% \vspace{-0.3cm}
\section{Conclusion}
We propose DGS-SVDD, a structured anomaly detection framework for cyber-physical systems using graph stream data. 
To this end, we integrate spatiotemporal patterns, modeling dynamic characteristics, deep representation learning, and one-class detection with SVDD. Transformer-based encoder-decoder architecture is used to preserve the temporal dependencies within a time segment. 
The temporal embedding and the predefined connectivity of the CPS are then used to generate weighted attributed graphs from which the fused spatiotemporal embedding is learned by a spatial embedding module.
A deep neural network, integrated with one-class SVDD is then used to group the normal data points in a hypersphere from the learnt representations. Finally, we conduct extensive experiments on the SWaT dataset to illustrate the superiority of our method as it delivers 35.87\% and 19.32\% improvement in F1-score and AUC respectively. For future work, we wish to integrate a connectivity learning policy into the transformer so that it just does not learn the temporal representation, rather it also models the dynamic influence among sensors. The code can be publicly accessed at https://github.com/ehtesam3154/dgs\_svdd.

\bibliographystyle{splncs04}
\bibliography{refs}
\end{document}